# Applying convolutional neural networks to extremely sparse image datasets using an image subdivision approach

Johan P. Boetker[*]

Department of Pharmacy, University of Copenhagen, Universitetsparken 2, 2100 Copenhagen, Denmark

**Corresponding Author:** Johan P. Boetker (address: Universitetsparken 2, 2100 Copenhagen, Denmark, phone: +45 35 33 62 86, email: johan.botker@sund.ku.dk)

# 1. Abstract

Purpose: The aim of this work is to demonstrate that convolutional neural networks (CNN) can be applied to extremely sparse image libraries by subdivision of the original image datasets.

Methods: Image datasets from a conventional digital camera was created and scanning electron microscopy (SEM) measurements were obtained from the literature. The image datasets were subdivided and CNN models were trained on parts of the subdivided datasets.

Results: The CNN models were capable of analyzing extremely sparse image datasets by utilizing the proposed method of image subdivision. It was furthermore possible to provide a direct assessment of the various regions where a given API or appearance was predominant.

## 2. INTRODUCTION

The possibility for extracting information from images has within the image analysis field been sought for many decades (1). The concept of neural networks has also existed, at least since the world war II era, and is as such not an entirely new approach (2). These neural networks are however relying on the collection of training data and that this collection of data constitutes a sufficient amount. Otherwise, if the gathered amount of data is insufficient it may be impossible to train a neural network (3). Furthermore, the growing interest in neural network based learning is mainly driven by its superior performance in many image and video applications emerging today (4) so there is a need for its implementation in order to take advantage of its superior performance. However, many experiments conducted within the pharmaceutical field in particular and related scientific fields such as chemistry, biology and physics may not easily render a sufficient amount of training data without conveying a considerable or perhaps even unfeasible workload. It is therefore of interest to investigate options where neural networks can be employed in meaningful way on datasets where the number of obtained images is very low.

The aim of this work is therefore to demonstrate that convolutional neural networks can be applied to extremely sparse image datasets using an image subdivision approach.

# 3. MATERIAL AND METHODS

The experimental part of this work was performed utilizing mixtures of nitrofurantoin (Ph. Eur. quality, Unikem, Copenhagen, Denmark) and paracetamol (Ph. Eur. quality, Fagron, Copenhagen, Denmark), that was imaged using an iPhone 8 plus. Furthermore, data from Kundrat et al. (5) which was published under the Creative Commons Attribution (CC BY) (http://creativecommons.org/licenses/by/4.0/) was also utilized. All calculations were performed in MATLAB® R2017b (MathWorks, Natick, MA, US). The datasets were subdivided into subset datasets using a for loop script named 'Subdivide'. This 'Subdivide' effectively raster scans over the columns and the rows of the dataset. All the subset datasets were upon creation saved to new folders. The convolutional neural network (CNN) model and the training of the model was performed using the script 'Get_CNN'. The CNN algorithm is designed from the MathWorks deep learning network for classification (6). Finally prediction of new images was performed using the script 'Predict_with_CNN'. The utilized scripts can be retrieved from the data repository:

(https://sid.erda.dk/public/archives/a2bae59140f88399f732db2af91bc383/published-archive.html)

# 4. RESULTS AND DISCUSSION

## 4.1. Construction of the CNN model from conventional image data

The nitrofuranoin and paracetamol datasets can initially be subdivide into datasets of 20 pixels by 20 pixels (Fig. 1) and a CNN model can be constructed from the subdivided datasets of paracetamol and nitrofurantoin. An appropriate subdivided dataset size can be assessed iteratively by constructing CNN models of various sizes and then comparing the CNN models predicting capabilities. Furthermore, subsequent optimization of e.g. the number of CNN layers can also be performed utilizing such an iterative approach.

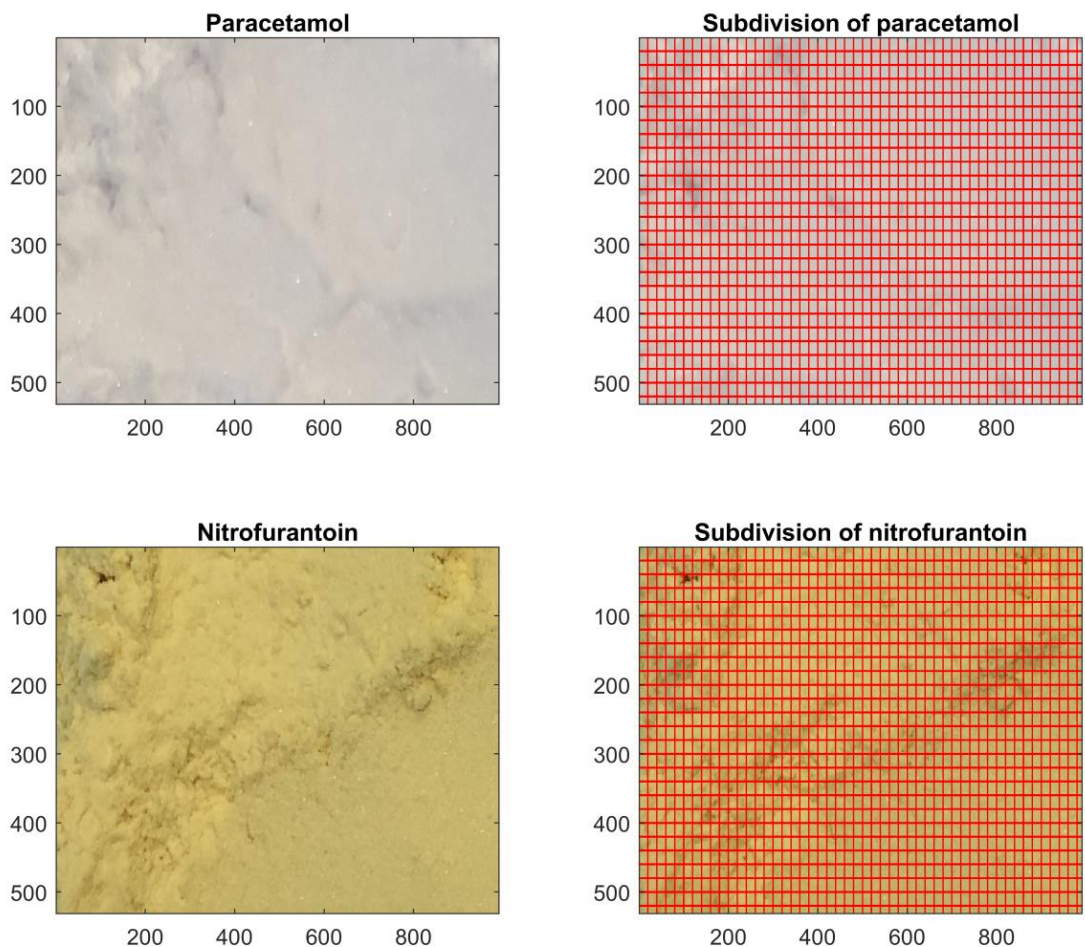

**Fig. 1** Paracetamol and nitrofurantoin conventional images and their subdivision.

The CNN model that has been constructed from the paracetamol subdivided dataset and the nitrofurantoin subdivided dataset can subsequently be utilized to predict the mix 1, mix 2, and mix 3 datasets (Fig. 2). From the second column in Fig. 2 it is directly observable that the CNN model is predicting that an increasing amount of

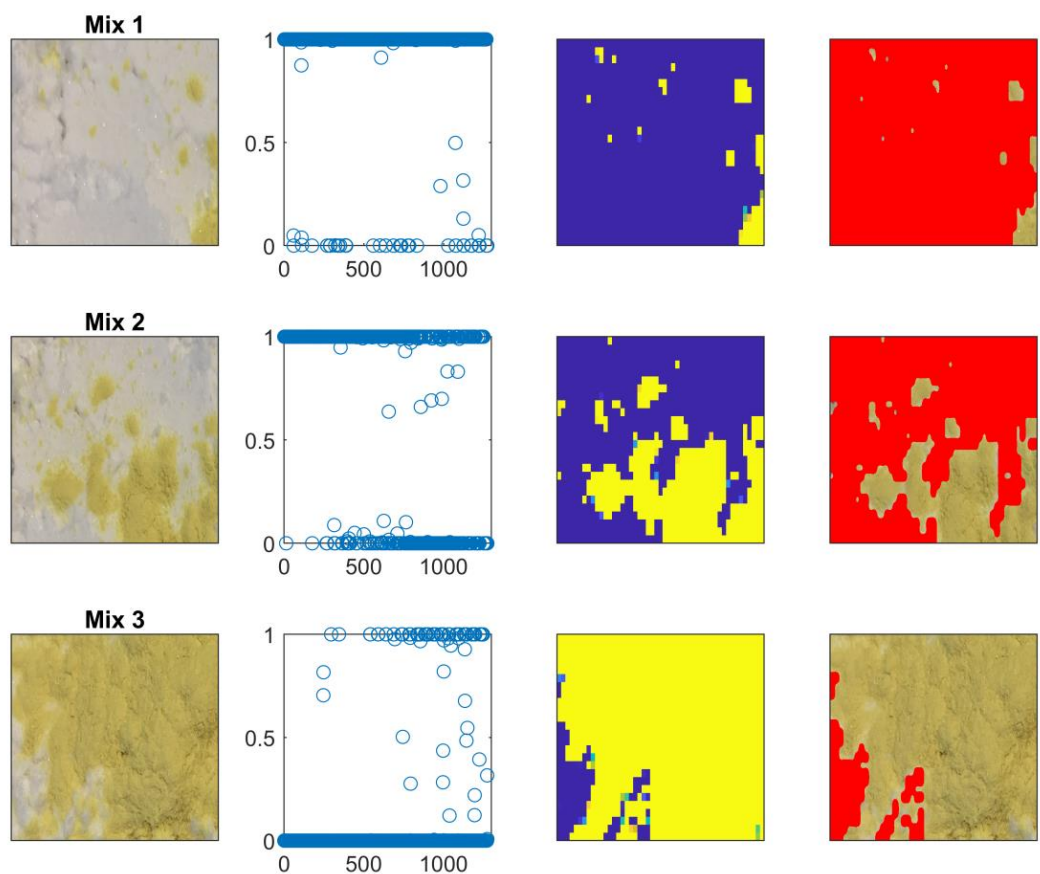

**Fig. 2** (first column) conventional images of mix 1, mix 2 and mix 3, (second column) prediction on the individual subdivided datasets where values close to one indicates paracetamol domains and values close to zero indicates nitrofurantoin domains, (third column), the subdivided dataset prediction values depicted in respect to spatial position where yellow indicates domains dominated by the nitrofurantoin and dark blue indicates domains dominated by paracetamol, (fourth column) a reshaped thresholded superimposed image where the predicted domains containing the paracetamol is colored red.

domains are successively dominated by nitrofurantoin as nitrofurantoin is added to the images. The data form the second column in Fig. 2 can be reshaped in respect to spatial position and colored according to the prediction values giving the third column in Fig. 2. From the third column in Fig. 2 a direct visualization of the various regions where either the paracetamol or the nitrofurantoin compound is predominant can be seen and this data can be further visualized by comparing the fourth column of Fig. 2, where a superimposed thresholded image is created masking out the paracetamol with red color, to the respective images presented in the first column of Fig. 2. It is hence evident that this method provides the capability of assessing whether a given image domain is dominated by a specific compound. In this case the distinction between paracetamol and nitrofurantoin is of course a trivial task due to the apparent color differences, but this trivialization is not a requirement as a convolutional neural network is capable of tackling more intricate relationships.

## 4.2. Construction of the CNN model from SEM data

The retrieved dataset from Kundrat et al. (5) displaying the morphological changes that occurred when different concentrations of poly(3-hydroxybuturyrate) was electrospun from a 1:1 ratio of chloroform and dichloromethane constitutes an interesting quantification challenge (Fig. 3).

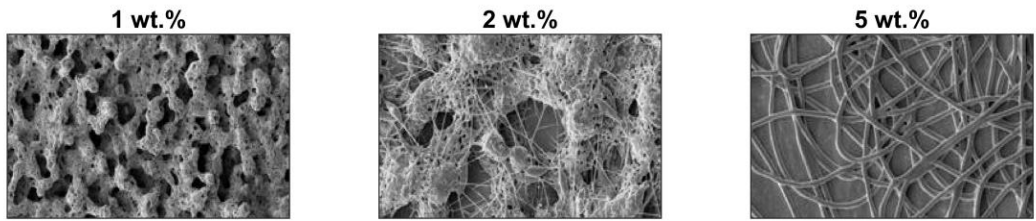

**Fig. 3** Morphological changes at varying concentrations (1 wt.%, 2 wt.% and 5 wt.%) of electrospun poly(3-hydroxybuturyrate) in a 1:1 ratio of chloroform and dichloromethane. Modified from (5) under the Creative Commons Attribution (CC BY) (http://creativecommons.org/licenses/by/4.0/).

A CNN model can be constructed from the left image that displays a porous morphology and the right image that displays a fiber like appearance. This obtained CNN model can subsequently be used to perform a prediction on the center image (Fig.3). Such a prediction that was based on a subset dataset sizes of 10 pixels by 10 pixels can be observed in Fig. 4 where it is directly apparent that this CNN model is capable of assessing which domains that are dominated by a fiber like appearance and which domains that are dominated by a porous appearance. This predicting capability is here directly linked to the morphology of the given domain as this is based on the grayscale SEM dataset that is devoid of any chemical information.

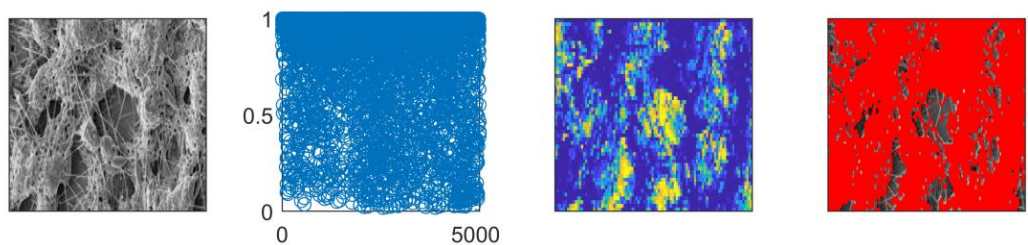

**Fig. 4** (first column) original SEM image modified from (5) under the Creative Commons Attribution (CC BY) (http://creativecommons.org/licenses/by/4.0/), (second column) prediction on the individual subdivided datasets where values close to one indicates porous domains and values close to zero fibrous domains, (third column), the subdivided dataset prediction values depicted in respect to spatial position where yellow indicates domains dominated by fibrous appearance and dark blue indicates domains dominated by porous appearance, (fourth column) a reshaped thresholded superimposed image where the predicted domains containing porous appearance is colored red.

These two datasets containing either conventional imaging data or SEM data have outlined a small subset of the possibilities that exist for utilizing CNN models with a subdivision approach. This method could be utilized for all imaginable spatial datasets where a starting point and/or an endpoint has been established. It is also directly possible to have additional classes so that multicomponent systems could be analyzed and e.g. background areas could be identified. Furthermore, this method could be utilized to search for various defects or artifacts within a dataset so it may have substantial applications under many different circumstances.

# 5. CONCLUSION

This study demonstrated the capability of applying convolutional neural networks with a subdivision approach to extremely sparse datasets. It was possible to detect different appearances in both conventional imaging data and SEM data so this method provides a possibility for evaluating domains based on the typical spectral appearance or morphology dominating a given particular domain. As this method is employing the adaptive capabilities of neural networks it is not necessarily constrained to only evaluate spectral or morphological appearances of compositions but may be utilized for whatever task that is at hand provided that the neural network and code are modified appropriately. Finally, it should also be mentioned that other machine learning methods such as support vector machines etc. can be directly incorporated utilizing this image subdivision approach.

# REFERENCES

1. Rosenfeld A. Iterative methods in image analysis. Pattern Recognition. 1978;10(3):181-187.
2. McCulloch WS, Pitts W. A logical calculus of the ideas immanent in nervous activity. The bulletin of mathematical biophysics. 1943;5(4):115-133.
3. Hammerstrom D. Working with neural networks. IEEE Spectrum. 1993;30(7):46-53.
4. Kuo CCJ. Understanding convolutional neural networks with a mathematical model. Journal of Visual Communication and Image Representation. 2016;41:406-413.
5. Kundrat V, Cernekova N, Kovalcik A, Enev V, Marova I. Drug Release Kinetics of Electrospun PHB Meshes. Materials (Basel). 2019;12(12).
6. MathWorks. Create Simple Deep Learning Network for Classification. https://www.mathworks.com/help/deeplearning/ug/create-simple-deep-learning-network-for-classification.html;jsessionid=523ab4f3a18318193be58542b76e Accessed 2020 AUG 25